\documentclass[10pt,twocolumn,letterpaper]{article}

\usepackage[pagenumbers]{cvpr} 

\usepackage[accsupp]{axessibility}  
\usepackage{graphicx}
\usepackage{amsmath}
\usepackage{amssymb}
\usepackage{booktabs}
\usepackage{multirow}
\usepackage{tabularx}
\usepackage[table]{xcolor}

\newcolumntype{C}{>{\centering\arraybackslash}X}
\usepackage[symbol]{footmisc}

\usepackage[pagebackref,breaklinks,colorlinks]{hyperref}

\usepackage[capitalize]{cleveref}
\crefname{section}{Sec.}{Secs.}
\Crefname{section}{Section}{Sections}
\Crefname{table}{Table}{Tables}
\crefname{table}{Tab.}{Tabs.}

\definecolor{3}{RGB}{255, 255, 200}
\definecolor{2}{RGB}{255, 220, 200}
\definecolor{1}{RGB}{255, 181, 163}
\definecolor{0}{RGB}{255, 255, 255}
\newcommand{\cc}[1]{\cellcolor{#1}}

\def\cvprPaperID{4953} 
\def\confName{CVPR}
\def\confYear{2023}

\begin{document}

\title{Local Implicit Ray Function for Generalizable Radiance Field Representation}

\author{Xin Huang$^{1}$\footnote[1]{}, Qi Zhang$^{2}$\footnote[2]{}, Ying Feng$^{2}$, Xiaoyu Li$^{2}$, Xuan Wang$^2$, Qing Wang$^1$\footnote[2]{} \vspace{6pt}\\
$^{1}$ School of Computer Science, Northwestern Polytechnical University, Xi'an 710072, China \\
$^{2}$ Tencent AI Lab
}


\twocolumn[{%
\renewcommand\twocolumn[1][]{#1}%
\maketitle

\begin{center}
    \centering
    \captionsetup{type=figure}
    \includegraphics[width=\hsize]{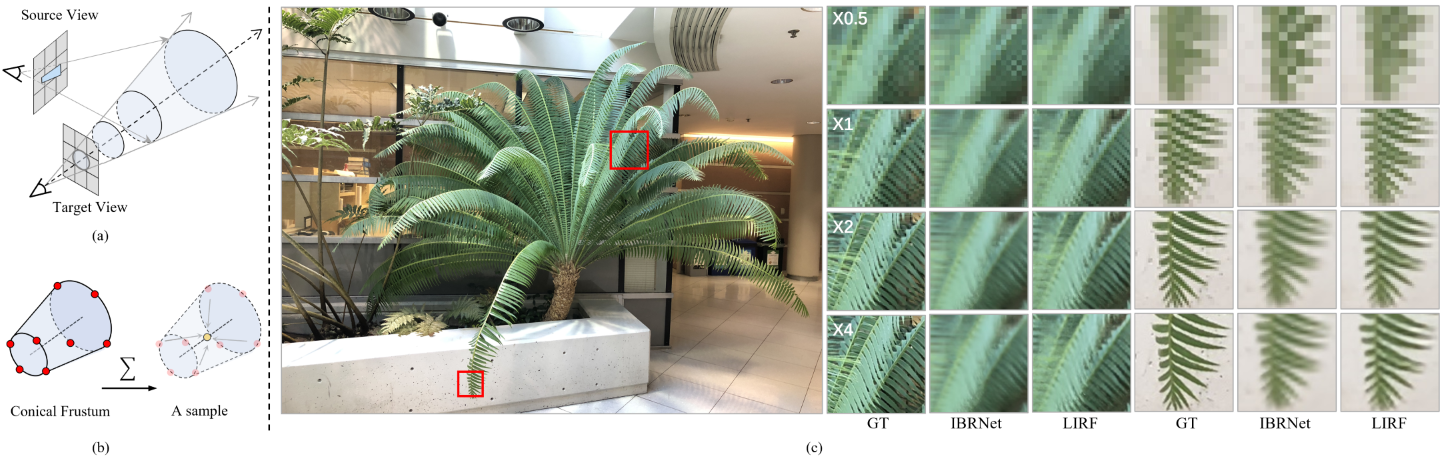}
    \vspace{-7mm}
    \caption{
    We propose LIRF to reconstruct radiance fields of unseen scenes for novel view synthesis. Given that current generalizable NeRF-like methods cast an infinitesimal ray to render a pixel at different scales, it causes excessive blurring and aliasing. Our method instead reasons about 3D conical frustums defined by the neighbor rays through the neighbor pixels (as shown in (a)). Our LIRF outputs the feature of any sample within the conical frustum in a continuous manner (as shown in (b)), which supports NeRF reconstruction at arbitrary scales. Compared with the previous method, our method can be generalized to represent the same unseen scene at multiple levels of details (as shown in (c)). Specifically, given a set of input views at a consistent image scale $\times1$, LIRF enables our method to both preserve sharp details in close-up shots (anti-blurring as shown in $\times2$ and $\times4$ results) and correctly render the zoomed-out images (anti-aliasing as shown in $\times0.5$ results).
    }  
    \label{fig:teaser}
\end{center}%
}]

\footnotetext[1]{Work was done during an internship at Tencent AI Lab.}
\footnotetext[2]{Corresponding authors.}

\begin{abstract}
    We propose \textbf{LIRF} (\textbf{L}ocal \textbf{I}mplicit \textbf{R}ay \textbf{F}unction), a generalizable neural rendering approach for novel view rendering. Current generalizable neural radiance fields (NeRF) methods sample a scene with a single ray per pixel and may therefore render blurred or aliased views when the input views and rendered views capture scene content with different resolutions. To solve this problem, we propose LIRF to aggregate the information from conical frustums to construct a ray. Given 3D positions within conical frustums, LIRF takes 3D coordinates and the features of conical frustums as inputs and predicts a local volumetric radiance field. Since the coordinates are continuous, LIRF renders high-quality novel views at a continuously-valued scale via volume rendering.  Besides, we predict the visible weights for each input view via transformer-based feature matching to improve the performance in occluded areas. Experimental results on real-world scenes validate that our method outperforms state-of-the-art methods on novel view rendering of unseen scenes at arbitrary scales. 
\end{abstract}

\section{Introduction}
\label{sec:intro}

Novel view synthesis has garnered recent attention with compelling applications of neural rendering in virtual and augmented reality. Different from image-based rendering \cite{levoy1996light, gortler1996lumigraph, buehler2001unstructured,zhou2018stereo, mildenhall2019local}, Neural Radiance Fields (NeRF) \cite{mildenhall2020nerf} implicitly represents the 3D scenes within multilayer perceptrons (MLPs) by mapping  coordinates to color and geometry of scenes. To render a pixel, the ray projected to that pixel is traced and the color of each sampled point along the ray is accumulated based on volume rendering. 

Despite NeRF and its variants having demonstrated remarkable performance in providing immersive experiences in various view synthesis tasks, their practical applications are constrained by the requirement of training from scratch on each new scene, which is time-consuming.
To overcome this problem, many researches \cite{yu2021pixelnerf,chibane2021stereo,wang2021ibrnet,li2021mine,chen2021mvsnerf,liu2022neural,johari2022geonerf,suhail2022generalizable} introduce image-based rendering techniques to NeRF, which achieves generalization on unseen scenes. They take into consideration the image features (from nearby views) of a 3D point. The common motivation is to predict the density and color of this point by matching the multi-view features, which is similar to the stereo matching methods \cite{schonberger2016pixelwise,yao2018mvsnet,gu2020cascade} that find a surface point by checking the consistency of multi-view features.

\begin{figure}[tb]
    \centering
    \includegraphics[width=\hsize]{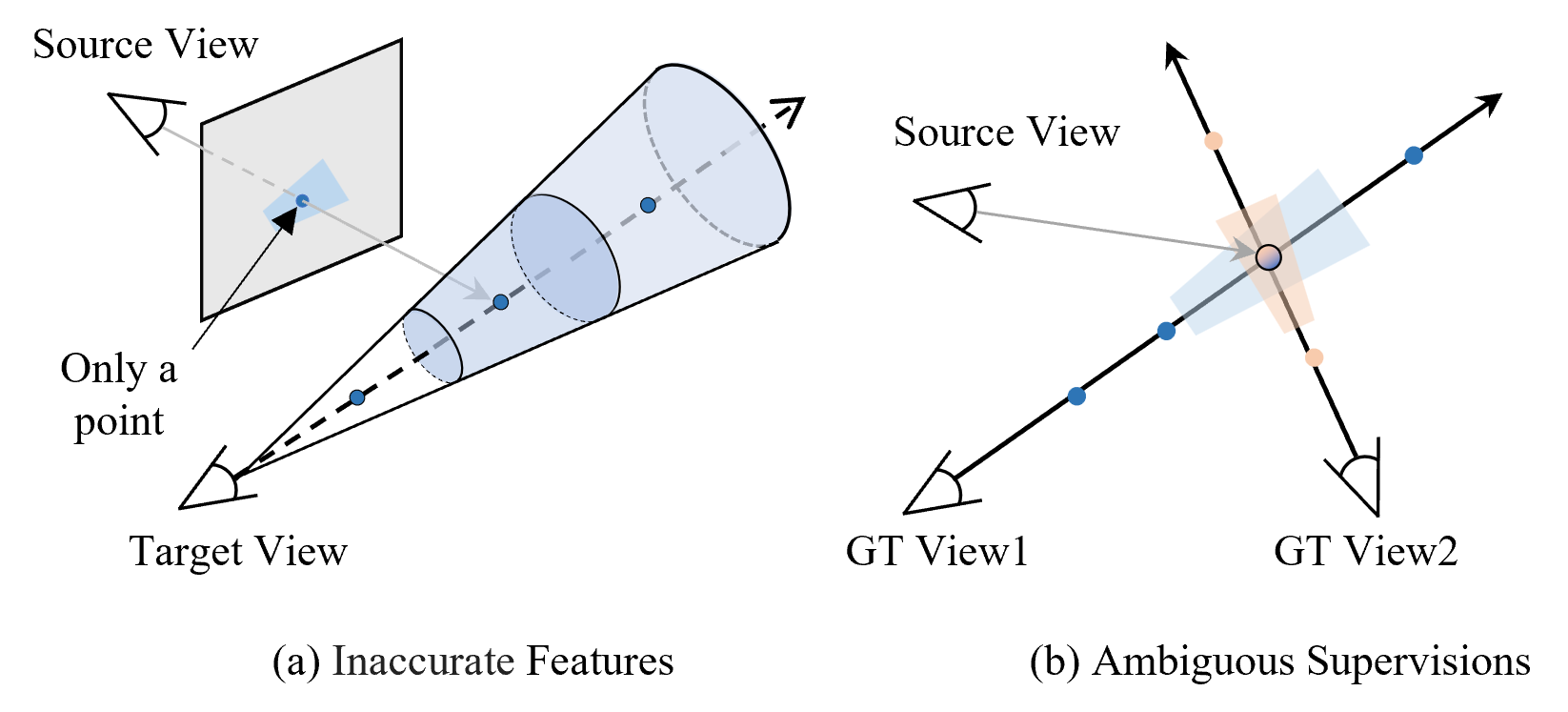}
    \vspace{-7mm}
    \caption{
    Most generalizable variants of NeRF represent a ray as a set of infinitesimal samples (shown here as dots) along that ray and map these samples into input views to query image features for volumetric representation prediction. However, this results in two drawbacks when training on multi-scale images with less constrained settings: 
    (a) Inaccurate features. The sampling strategy which ignores the shape and size of each ray is difficult to query accurate image features. (b) Ambiguous supervisions. The same 3D position captured by cameras under different scales results in different colors because these pixels are the integral of regions with different shapes and sizes (shown here as trapezoids). During the training, the network learns to map the same image features (from the source view) to these different colors, which causes ambiguous results.}
    \label{fig:motivation}
    \vspace{-5mm}
\end{figure}

While these methods generalize well on new scenes when the distance of input and testing views are roughly constant from the scene (as in NeRF), they cannot properly deal with the less constrained settings such as different resolutions or varying focal length and produce results with blurring or aliasing artifacts. Since a single ray is cast through each pixel whose size and shape are ignored, it's challenging to query the accurate feature of the target ray from input images as shown in \cref{fig:motivation}(a), and the model learns an ambiguous result as shown in \cref{fig:motivation}(b). Mip-NeRF \cite{barron2021mip}, a NeRF variant of per-scene optimization, proposes an anti-aliasing design that models the ray through a pixel as a cone and uses a 3D Gaussian to approximate the sampled \textit{conical frustum} (a cone cut perpendicular to its axis) for volumetric representation. However, directly extending Mip-NeRF to a generalizable method is also challenging to extract the accurate features of the ray from input images due to the subpixel precision. Consequently, an efficient solution is to supersample each pixel by marching multiple rays according to its footprint, similar to the strategy used in offline raytracing.

Our key insight is the \emph{local implicit ray function} (LIRF) that represents the feature aggregation of samples within ray conical frustum in a continuous manner, as shown in \cref{fig:teaser}. 
Specifically, given any 3D sampled position within a conical frustum, our LIRF outputs the aggregated feature by taking the 3D coordinate of this position and the features of vertices within the conical frustum as inputs (the vertices of a conical frustum are defined with eight points (red points) as shown in \cref{fig:teaser}). The continuous sampled position allows our method to arbitrarily upsample the rendered rays and thus synthesize novel views of the same unseen scene at multiple levels of detail (anti-blurring and anti-aliasing).
Furthermore, recent generalizable NeRF methods \cite{liu2022neural,johari2022geonerf} introduce multi-view depth estimation to reduce the artifacts caused by occlusions, but it is computationally expensive to construct the cost volume for each view. We instead match local multi-view feature patches to estimate the visibility weights of each sample for anti-occlusion. Overall, our main contributions are:
\vspace{-2mm}
\begin{enumerate}
    \setlength\itemsep{0pt}
    \item A new generalizable approach that renders pixels by casting cones and outperforms existing methods on novel view synthesis at multiple scales.
    \item A local implicit ray function that simplifies the representation of conical frustums and enables continuous supersampling of rays.  
    \item A transformer-based visibility weight estimation module that alleviates the occlusion problem. 
\end{enumerate}
To evaluate our method, we construct extensive series of experiments on real forward-facing scenes. Our experiments show that trained on large amounts of multi-view data, LIRF outperforms state-of-the-art generalizable NeRF-like methods on novel views synthesis for unseen scenes. 

\section{Related Work}

\noindent\textbf{Image-based rendering.} When the images for a scene are captured densely, earlier lines of work directly interpolate novel views from input views \cite{levoy1996light, gortler1996lumigraph} or render views by weighted blending \cite{buehler2001unstructured, davis2012unstructured}. More recently, researchers have been focused on the issue of novel views synthesis from sparse inputs. To map and blend the input images in a novel target view, some methods \cite{cayon2015bayesian, overbeck2018system} use the proxy geometry obtained via structure-from-motion (SfM) or multi-view stereo (MVS). Moreover, to improve the accuracy of mapping and blending, some improved methods are proposed for proxy geometry estimation \cite{chaurasia2013depth, hedman2016scalable}, optical flow correction \cite{eisemann2008floating,du2018montage4d} and deep blending \cite{hedman2018deep, riegler2020free, riegler2021stable}. Differently, some other methods directly reconstruct the texture of mesh \cite{debevec1998efficient,thies2019deferred,huang2020adversarial} or points cloud \cite{meshry2019neural,pittaluga2019revealing,aliev2020neural,ruckert2022adop} for novel view synthesis. However, the rendering quality by IBR methods is directly affected by geometric inaccuracies of 3D reconstruction methods \cite{jancosek2011multi,schonberger2016structure}, especially in low-textured or reflective regions. 

\noindent\textbf{Explicit volumetric representations.} The other methods render high-quality novel views by constructing an explicit scene representation, such as voxel grids \cite{tulsiani2017multi, kar2017learning} and layered-depth images \cite{shade1998layered,penner2017soft}, from captured images. For example, Zhou \textit{et al} \cite{zhou2018stereo}. represent a scene with a set of depth-dependent images with transparency, which is called multi-plane images (MPIs). Novel views can be directly rendered from MPIs using alpha compositing. Recently, MPI representation has been applied in many scenarios, such as rendering novel views from multiple MPIs \cite{mildenhall2019local}, novel view synthesis from a single image \cite{tucker2020single, li2020synthesizing}, variants as multi-sphere images \cite{broxton2020immersive, attal2020matryodshka}, and 3D image generation \cite{zhao2022generative,xiang2022gram}. Trained on a large multi-view dataset, MPI-based methods can generalize well to unseen scenes and render photo-realistic novel views fast. However, those methods always render views within limited viewing volumes. Besides, the volumetric representations explicitly decompose a scene into extensive samples, which requires large memory to store them and limits the resolution of novel views.

\noindent\textbf{Neural scene representations.} Representing the geometry and appearance of a scene with neural networks has been a surge. Traditional methods explicitly represent scenes with point clouds \cite{wu2020multi,ruckert2022adop}, meshes \cite{thies2019deferred,huang2020adversarial}, or voxels \cite{lombardi2019neural, sitzmann2019deepvoxels}, while neural scene representations implicitly encode the scene with continuous coordinate-based functions such as radiance fields \cite{mildenhall2020nerf,barron2021mip}, signed distance fields \cite{chabra2020deep,genova2020local,yariv2021volume}, or occupancy fields\cite{mescheder2019occupancy,peng2020convolutional}. NeRF \cite{mildenhall2020nerf} approximates a 3D scene with an implicit function and has shown high-quality view synthesis results. Mip-NeRF \cite{barron2021mip} reduces objectionable aliasing artifacts and improves NeRF’s ability to represent scene details by casting a cone instead of a ray. While NeRF has been expanded to many new scenarios\cite{boss2021nerd,srinivasan2021nerv,li2021neural,xian2021space,pumarola2021d,guo2020object,zhang2021editable,huang2022hdr,ma2022deblur,chen2022hallucinated}, most NeRF-like models take coordinate as inputs, which restricts their ability to generalize to unseen scenes. Recently, some generalizable NeRF-like methods have been proposed\cite{yu2021pixelnerf,wang2021ibrnet,chibane2021stereo,li2021mine,chen2021mvsnerf,liu2022neural,johari2022geonerf}. PixelNeRF \cite{yu2021pixelnerf}, SRF \cite{chibane2021stereo}, MINE \cite{li2021mine} and MVSNeRF \cite{chen2021mvsnerf} try to construct radiance fields on-the-fly from sparse input views, while they struggle with complex scenes. To solve occlusions, NeuRay \cite{liu2022neural} and GeoNeRF \cite{johari2022geonerf} estimate depth using MVS methods \cite{yao2018mvsnet,gu2020cascade} and calculate the occlusion masks from the estimated depth maps. However, it's expensive to construct cost volumes per source view to estimate depth. While these generalizable methods have been capable of rendering novel views for unseen scenes, rendering novel views by casting a single ray may produce renderings that are aliased or blurred. 

\begin{figure*}[tb]
    \centering
    \includegraphics[width=\hsize]{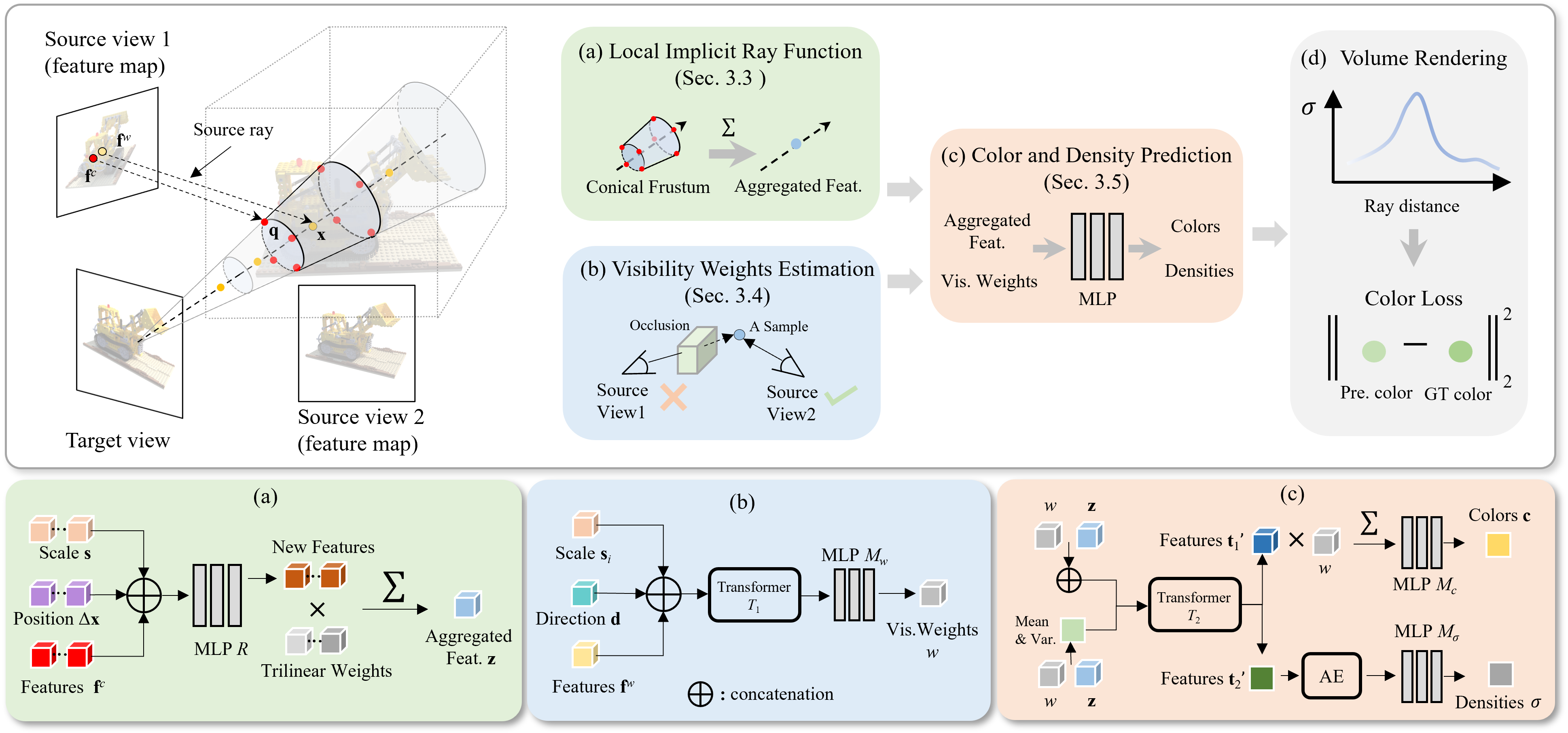}
    \vspace{-6mm} 
    \caption{
    The overview of LIRF. To render a target image, a set of $V$ neighboring source views are selected and the 2D feature map of each source view is extracted via an EDSR network \cite{lim2017enhanced}. $N$ points (yellow points) are sampled along a target ray, and the corresponding conical frustum is represented with $M$ vertices (red points). (a) For a sample on target ray at 3D position $\mathbf{x}$, we obtain its feature $\mathbf{z}$ by aggregating the features of the vertices and two latent codes (relative position $\Delta \mathbf{x}=\mathbf{x}-\mathbf{q}$ and target scale $\mathbf{s}$). (b) To deal with occlusions, we also estimate the visibility weights of each source view. Visibility weights can be considered the probability of a sample being observed by each source view. Taking the image features $\mathbf{f}^w$, source ray direction $\mathbf{d}$ and target scale $\mathbf{s}$ as inputs, our model outputs the visibility weights in source views of sample $\mathbf{x}$. (c) After obtaining the features $\mathbf{z}$ and corresponding visibility weights $w$, we aggregate them by a transformer \cite{vaswani2017attention} module $\mathcal{T}_2$ which outputs two parts of features: one for the colors and the other for densities. Next, we explicitly blend features ${\mathbf{t}_1}'$ and map them into colors $\mathbf{c}$ by an MLP $\mathcal{M}_{c}$. As for features ${\mathbf{t}_2}'$, an AE network \cite{johari2022geonerf} aggregates their information along a ray and an MLP $\mathcal{M}_{\sigma}$ maps the features to densities $\sigma$. (d) Finally, the color of target ray is rendered by volume rendering from the colors $\mathbf{c}$ and densities $\sigma$. The mean squared error between the predicted color and ground truth color is calculated for optimization.
    }
    \label{fig:pipeline}
    \vspace{-3mm} 
\end{figure*}

\section{Method}
Our goal is to predict volumetric radiance fields from a set of multi-view images captured at a consistent image scale ($\times 1$), and output novel views at continuous scales ($\times0.5 \sim \times4$). Our proposed framework is presented in \cref{fig:pipeline}, which is composed of five parts: 1) extracting 2D feature maps from source images (\cref{sec:feature}), 2) obtaining the image feature for the samples on target rays via local implicit ray function (\cref{sec:ray_function}), 3) predicting the visibility weights of each source view by matching feature patches (\cref{sec:occlusion}), 4) aggregating local image features from different source views and mapping them into colors and densities (\cref{sec:color_density}), 5) rendering a target pixel via volume rendering (\cref{sec:volume_rendering}). 

\subsection{Volume Rendering} \label{sec:volume_rendering}
We first briefly review the volume rendering in NeRF \cite{mildenhall2020nerf}.
NeRF implicitly encodes a scene as a continuous radiance field. Given a ray $\mathbf{r}$ passing through the scene, NeRF maps the 5D coordinates (3D for position, 2D for direction) of each sampled location into view-dependent color and volume density via an MLP. To determine the pixel color of the ray $\mathbf{r}$, NeRF first obtains the colors and densities of all samples on the ray $\mathbf{r}$ and then accumulates the colors according to densities. The volume rendering thus is defined by:
\begin{equation}\small
\begin{split}
	\widehat{C}(\mathbf{r}) & =\sum_{i=1}^{N}T_i(1-\mathrm{exp}(-\sigma_i\delta_i))\mathbf{c}_i, \\
	T_i & = \mathrm{exp}\left( - \sum_{l=1}^{i-1}\sigma_l \delta_l \right),
	\label{eq:volume_rendering}
\end{split}
\end{equation}
where $N$ is the number of samples along a ray with ascending depth values. The $\mathbf{c}$ and $\sigma$ denote colors and densities, respectively. $T$ denotes the accumulated transmittance. The $\delta$ is the distance between adjacent samples. Following IBRNet \cite{wang2021ibrnet}, we remove $\delta$ in volume rendering for a better generalization. To render a novel view of unseen scenes, we will introduce our proposed method to obtain colors and densities from input images instead of coordinates.

\subsection{Image Feature Extraction} \label{sec:feature}
Given a target viewpoint, we select a set of $V$ nearby source images and their corresponding camera poses as inputs. Our method relies on the local image features from source images to produce target images. Usually, most generalizable methods \cite{wang2021ibrnet, liu2022neural} extract dense image features from each source view by a U-Net \cite{ronneberger2015u} architecture with ResNet \cite{he2016deep}. Although U-Net and ResNet are widely adopted in high-level computer vision tasks such as image recognition and image segmentation, our rendering is based on pixel-wise operations. The feature extraction network is supposed to pay more attention to image details, so we use an EDSR network \cite{lim2017enhanced} to extract image features. The EDSR network removes unnecessary modules in conventional residual networks, which makes the network more focused on detailed content. We modify the last layer in the EDSR network to output two image features for the following modules, with one image feature for the estimation of visibility weights, and the other for the prediction of colors and densities. Given a source image $I$, the image feature extraction is formulated as:
\begin{equation}\small
	(\mathbf{F}^{w}, \mathbf{F}^{c})=\mathrm{EDSR}(I),
	\label{eq:image_feature}
\end{equation}
where $\mathbf{F}^{w}$ denotes the feature map for visibility weights and  $\mathbf{F}^{c}$ denotes the feature map for colors and densities.

\subsection{Local Implicit Ray Function} \label{sec:ray_function}
In our method, a target ray is defined as the integral of a conical volume. To consider the shape and size of the volume viewed by each ray, we define the feature of each sample along the ray by a continuous integration within the corresponding conical frustum. Given the 3D position $\mathbf{x}$ within the conical frustum, it's feature $\mathbf{z}$ is defined as:
\begin{equation}\small
	\mathbf{z}=\int W(|| \Delta \mathbf{x}_\mathbf{q} ||_2) \mathcal{R}(\mathbf{f}_q, \Delta \mathbf{x}_\mathbf{q}) \ \mathrm{d}\mathbf{q},
	\label{eq:ray_function1}
\end{equation}
where $\mathbf{q}$ denotes a 3D position within the conical frustum, $\mathbf{f}_\mathbf{q}$ is the feature at position $\mathbf{q}$, $\Delta \mathbf{x}_\mathbf{q}=\mathbf{x}-\mathbf{q}$ denotes the relative position between two points, and $||\cdot||_2$ denotes the Euclidean distance.  $W$ is a weights function that outputs weights according to the distance $|| \Delta \mathbf{x}_\mathbf{q} ||_2$. $\mathcal{R}$ is an MLP-based function for the aggregation of features and positions.

\noindent\textbf{Discrete representation.} Considering the computation and memory consumption, we cast four rays from the cone’s footprint to discretely approximate the continuous conical volume. Inspired by Plenoxels \cite{fridovich2022plenoxels}, a voxel-based representation, each conical frustum along the cone is represented with 8 vertices, and the target ray passes through all conical frustums, as shown in \cref{fig:method}(b). Any samples within the conical frustum can be obtained by trilinear interpolation, which is similar to the sampling strategy of voxel-based representation. Specifically, we project the vertices on the four rays into the source views and query corresponding features $\mathbf{f}^{c}$ from feature maps $\mathbf{F}^{c}$. For a sample at position $\mathbf{x}$ within the corresponding conical frustum, its feature is defined as:
\begin{equation}\small
\begin{split}
	&\mathbf{z} = \sum_{j=1}^{M} W(||\Delta \mathbf{x}_{j}||_2)\mathcal{R}(\mathbf{f}_{j}^{c}, \Delta \mathbf{x}_{j}), \\
	&W(||\Delta \mathbf{x}_{j}||_2) = \frac{||\Delta \mathbf{x}_{j}||_2 }{\sum_{l=1}^{M} ||\Delta \mathbf{x}_{l}||_2},
	\label{eq:ray_function2}
\end{split}
\end{equation}
where $M=8$ is the number of vertices used to represent a conical frustum. $\mathbf{f}^{c}$ is the image feature of the vertices on the conical frustum. $\Delta\mathbf{x}$ denotes the relative position between the sample and the vertices within the conical frustum.

\noindent\textbf{Cone radius.} A cone cast from the view origin and passing through the image plane, the radius of the cone at the image plane is called cone radius and parameterized as $r$. The cone radius is affected by image resolutions and observation distances (being closer to objects means a larger scale). However, it is hard to know the observation distances beforehand. Besides, since our testing scenes are captured using a hand-held camera, the observation distance of novel views and source views are slightly different. To adjust the cone size accurately and conveniently, we don't directly modify the cone size only according to the relative image resolution. Instead, we set a maximum cone radius $r_{m}$ which is the pixel width of the target image with the minimum resolution, and a latent code is introduced to implicitly control the size of a cone. Specifically, we first cast four rays passing through the four red points in the target image plane, as shown in \cref{fig:method}(a). For any target ray projecting within the circle with radius $r_{m}$, its features are aggregated from the features of the four neighboring rays. Moreover, a scale-dependent input $\mathbf{s}$ is additionally introduced to modify the cone size, where $\mathbf{s}$ is the relative scale of a target image. 
We reformulate \cref{eq:ray_function2} with the form:
\begin{equation}\small
	\mathbf{z}= \sum_{j=1}^{M} W(||\Delta \mathbf{x}_{j}||_2)
	\mathcal{R}(\mathbf{f}_{j}^{c}, \Delta \mathbf{x}_{j}, \mathbf{s}). 
	\label{eq:ray_function3}
\end{equation}
In practice, the positional encoding strategy \cite{mildenhall2020nerf} is applied on the relative position $\Delta \mathbf{x}$ but not on relative scale $\mathbf{s}$, since we want the implicit function to learn more high-frequency information of image feature and adjust the scale of renderings smoothly.

\subsection{Visibility Weights Estimation} \label{sec:occlusion}
\begin{figure}[tb]
    \centering
    \includegraphics[width=\hsize]{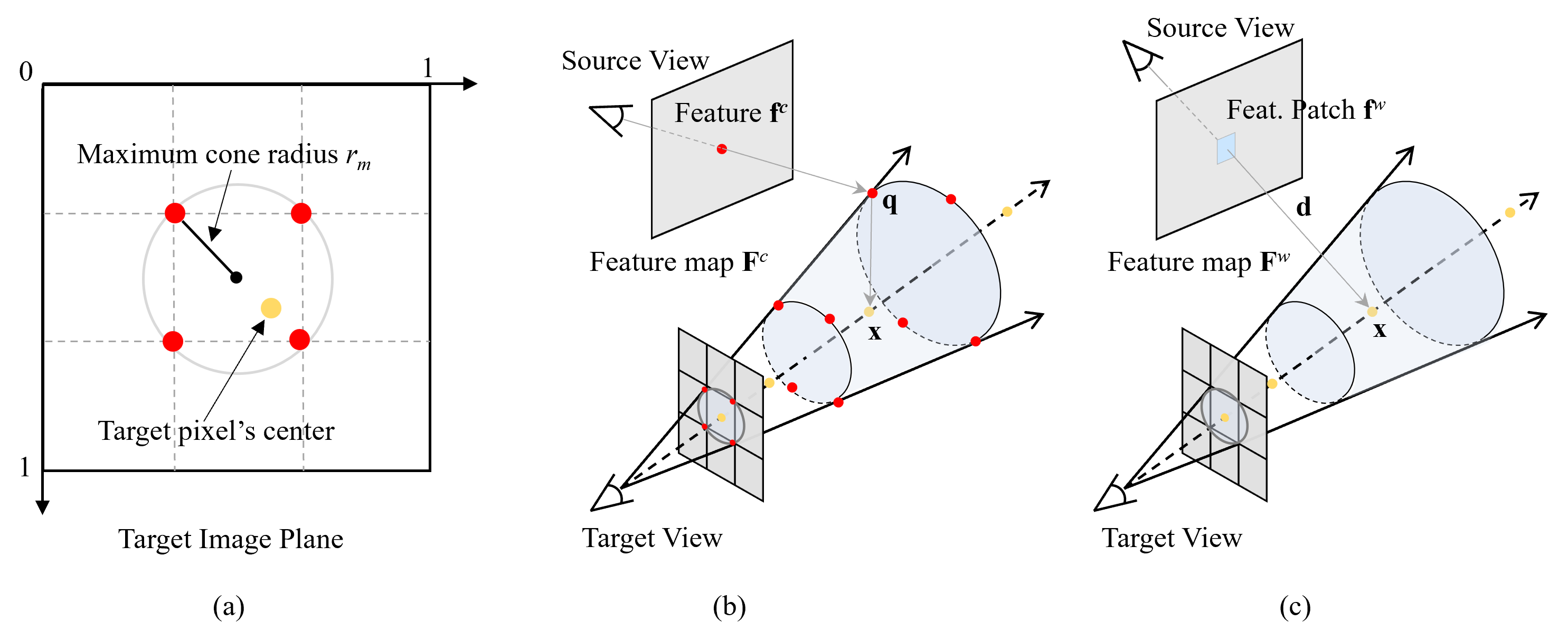}
    \vspace{-7mm}   
    \caption{Method details of LIRF. The $\mathbf{x}$ denotes the position of a sample, $\mathbf{q}$ denotes the position of a vertex, and $\mathbf{d}$ denotes the direction of a source ray. (a) The rays projecting at target image plane. The size of target image is normalized. (b) A cone is represented with four rays (solid lines, only two are drawn) and  a conical frustum is represented with eight vertices (red points). A target ray passes through the cone (dotted line). (c) The feature patches for visibility weights are obtained by warping samples on the target ray to feature maps.}
    \label{fig:method}
    \vspace{-3mm} 
\end{figure}

In this stage, we estimate the visibility weight that reveals the occlusions in each source view. Without any geometric priors, it's challenging to solve the occlusions thoroughly. As a compromise, we assume that the occluded content appears in most source views. Our model estimates the visibility weights of one source view by matching its feature patch with the feature patches from other source views. For the samples on a target ray, we obtain their feature patches $\mathbf{f}^{w}$ from the image features $\mathbf{F}^{w}$ as shown in \cref{fig:method}(c). Apart from the feature patches, two latent codes are introduced to improve the performance. To consider the direction for each source ray, their directions $\mathbf{d}$ in target camera coordinates are input. We do not use the global directions of source rays in world coordinates, since different scenes have different world coordinates and local directions are more suitable for the generalization to unseen scenes. To consider the scale of the target image, we also introduce the target scale $\mathbf{s}$. We define the feature concatenation of all inputs as:
\begin{equation}\small
	\mathbf{t}_0= \mathrm{MLP}( \mathbf{f}^{w} \parallel \mathbf{d} \parallel \mathbf{s} ),
	\label{eq:token1}
\end{equation}
where $\parallel$ denotes the concatenation operation and \textrm{MLP} denotes a two-layer MLP used to reduce the channels of features. Positional encoding \cite{mildenhall2020nerf} is applied on directions $\mathbf{d}$.

Next, a self-attention transformer \cite{vaswani2017attention} module $\mathcal{T}_1$ is used to fully match these aggregated feature $\mathbf{t}_0$  and output new features for visibility weights. Formally, this stage is written as:
\begin{equation}\small
	{\mathbf{t}_0}'= \mathcal{T}_1 (\mathbf{t}_0).
	\label{eq:t1}
\end{equation}
We then map the new feature ${\mathbf{t}_0}'$ into visibility weights $w$ by an MLP $\mathcal{M}_w$,
\begin{equation}\small
	w = \mathcal{M}_w ( {\mathbf{t}_0}' ).
	\label{eq:mw}
\end{equation}

\subsection{Color and Density Prediction} \label{sec:color_density}
After obtaining the local image features $\mathbf{z}$ and the visibility weights $w$, the colors and densities can be predicted. We aggregate the features and visibility weights to features $\mathbf{t}_1$. Besides, the weighted mean and weighted variance of features $\mathbf{z}$ over all source views are calculated based on visibility weights $w$. The mean and variance are then aggregated into features $\mathbf{t}_2$. This process is defined as:
\begin{equation}\small
\begin{split}
	\mathbf{t}_1 & = \mathrm{MLP}( \mathbf{z} \parallel w ), \\
	\mathbf{t}_2 & = \mathrm{MLP}( mean(\mathbf{z}, w ) \parallel var(\mathbf{z}, w) ).
	\label{eq:token2}
\end{split}
\end{equation}
$\mathbf{t}_1$ could be considered as the color information from source views, while $\mathbf{t}_2$ could be considered as the measure of image feature consistency. The densities are predicted by checking the feature consistency, since local image features from different source views are usually consistent when the sample is on the object surface \cite{liu2022neural}.

Next, the two features are fed into a self-attention transformer $\mathcal{T}_2$ for fully information aggregation,
\begin{equation}\small
	({\mathbf{t}_1}',  {\mathbf{t}_2}' )  = \mathcal{T}_2 ( \mathbf{t}_1,  \mathbf{t}_2).
	\label{eq:t2}
\end{equation}
The feature ${\mathbf{t}_1}'$ and ${\mathbf{t}_2}'$ now have combined the information from the the local image features and visibility weights. Similar to most generalizable NeRF-like methods \cite{wang2021ibrnet,liu2022neural,johari2022geonerf}, the density is estimated by an auto-encoder-style network \cite{johari2022geonerf} $\mathrm{AE}$ which aggregates information along a ray, and an MLP $\mathcal{M}_{\sigma}$ that maps features to densities,
\begin{equation}\small
	\sigma = \mathcal{M}_\sigma \left( \mathrm{AE}({\mathbf{t}_2}') \right).
	\label{eq:ms}
\end{equation}

As for colors, we further explicitly blend the features ${\mathbf{t}_1}'$ according to the visibility weights, and then an MLP $\mathcal{M}_c$ are used for color prediction,
\begin{equation}\small
	\mathbf{c} = \mathcal{M}_c \left( \sum_{k=1}^V ({\mathbf{t}_1}'_{,k} w_{k}) \right).
	\label{eq:mc}
\end{equation} 
Once the densities and colors of the samples on target rays are predicted, the final colors of target rays are rendered via volume rendering (\cref{eq:volume_rendering}).

\subsection{Loss Function}
We minimize the mean squared error between our predicted colors and ground truth colors for optimization:
\begin{equation}\small
	\mathcal{L} \!=\! \sum_{\mathbf{r}\in{\Omega}} \| \widehat{C}(\mathbf{r}) - C(\mathbf{r}) \|^2_2,
	\label{eq:loss_color}
\end{equation}
where $\Omega$ is a set of camera rays at target position. $\widehat{C}$ and $C$ are the predicted colors and ground truth colors, respectively.

\begin{figure*}[tb]
    \centering
    \includegraphics[width=\hsize]{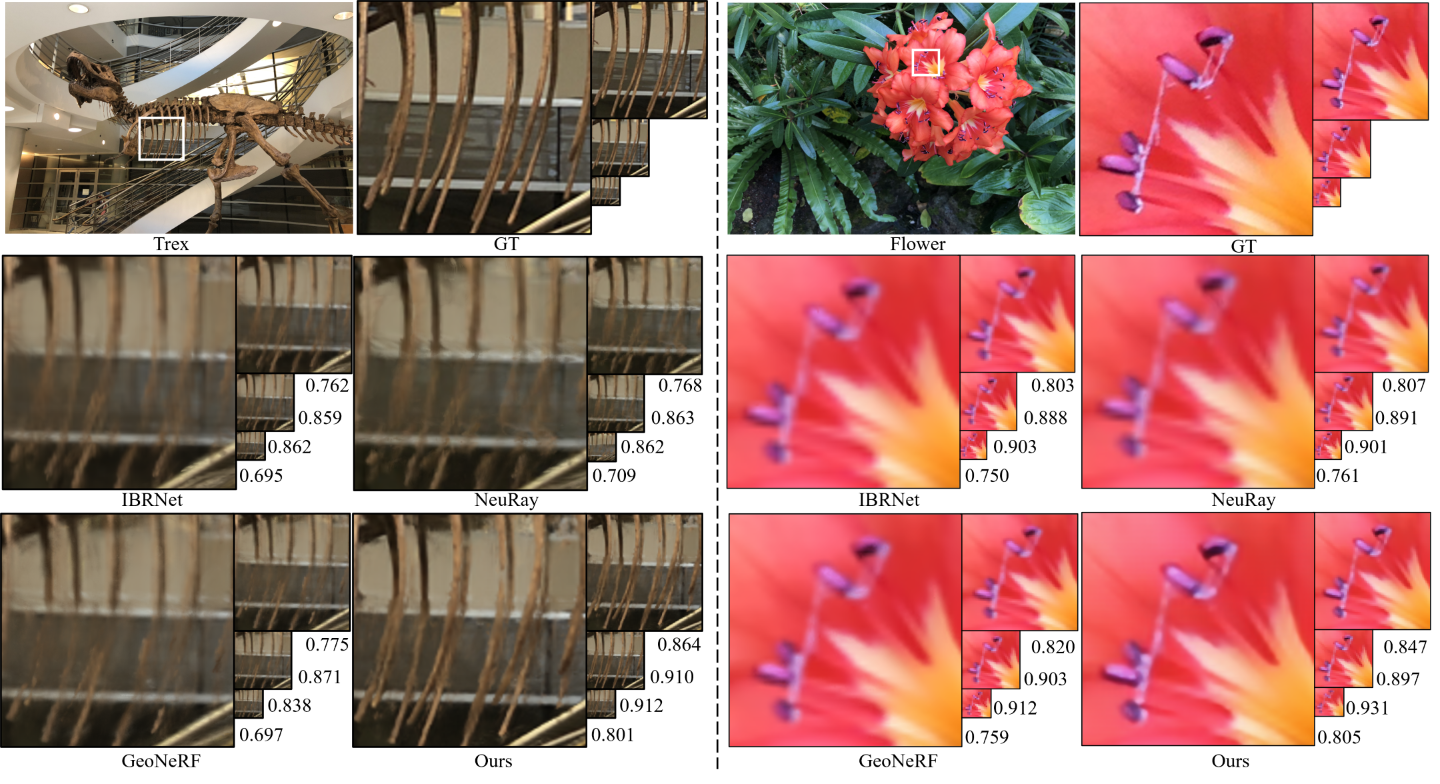}
    \vspace{-7mm}
    \caption{Qualitative comparisons on real forward-facing scenes. The cropped regions of two scenes are rendered at different scales ($\times0.5$, $\times1$,$\times2$ and $\times4$). SSIM averaged over all novel views for each scale are provided at the lower right of images.  }
    \label{fig:results}
\end{figure*}

\section{Experiments}

\subsection{Implementation Details} 
To render a target view, we select nearby source views using $V=8$ during both training and testing. $M=8$ vertices are used to represent a conical frustum. The cone radius $r_m$ is set to the pixel width of the target image with the minimum resolution. We directly sample $N=128$ points on a ray instead of using the coarse-to-fine sampling \cite{mildenhall2020nerf}. The extracted image feature maps have $32$ channels and their size is the same as the original image size. The size of feature patches used for visibility weights is set to $7\times7$.
We train the generalizable LIRF for 250k iterations. The training batch size of rays is set to $512$. We use Adam optimizer \cite{kingma2014adam} with a learning rate of $5\times10^{-4}$ that decays exponentially along with the optimization. We train our model on eight V$100$ GPUs, which takes about two days.

\begin{table*}[!tb]
\small
\centering
\caption{Quantitative comparisons of LIRF and its ablations against IBRNet \cite{wang2021ibrnet}, NeuRay \cite{liu2022neural} and GeoNeRF \cite{johari2022geonerf} on LLFF multi-scale testing dataset. Metrics are averaged over all scenes. $*$ denotes training on the same multi-scale training set as our method.}
\begin{tabularx}{\linewidth}{@{}l|CCCC|CCCC|CCCC|C}
\hline
       & \multicolumn{4}{c}{PSNR$\uparrow$} & \multicolumn{4}{|c}{SSIM$\uparrow$}  & \multicolumn{4}{|c|}{LPIPS$\downarrow$} & \multirow{2}{*}{Avg.$\downarrow$}\\
         \cline{2-5} \cline{6-9} \cline{10-13} 
       & $\times0.5$  & $\times1$ & $\times2$  & $\times4$  & $\times0.5$  & $\times1$ & $\times2$  & $\times4$ & $\times0.5$  & $\times1$ & $\times2$  & $\times4$  &\\
\hline
IBRNet      & 25.06 & 25.28 & 23.18 & 22.15 & 0.866 & 0.840 & 0.731 & 0.669 & 0.108 & 0.160 & 0.299 & 0.442 & 0.079 \\
NeuRay      & 24.80 & 25.17 & 23.10 & 22.08 & 0.859 & 0.837 & 0.729 & 0.667 & 0.107 & 0.157 & 0.294 & 0.434 & 0.080 \\
GeoNeRF     & 24.89 & 25.74 & 23.43 & 22.27 & 0.864 & 0.864 & 0.753 & 0.679 & 0.108 & 0.136 & 0.274 & 0.421 & 0.076 \\
\hline
IBRNet*     & 22.96 & 23.62 & 22.33 & 21.51 & 0.816 & 0.813 & 0.723 & 0.665 & 0.140 & 0.178 & 0.307 & 0.444 & 0.090 \\
NeuRay*     & 22.79 & 22.40 & 21.23 & 20.61 & 0.794 & 0.733 & 0.646 & 0.622 & 0.172 & 0.262 & 0.382 & 0.493 & 0.107 \\
GeoNeRF*    & 23.39 & 25.08 & 23.81 & 22.69 & 0.821 & 0.859 & 0.784 & 0.708 & 0.138 & 0.134 & 0.255 & 0.401 & 0.077 \\
\hline
Ours                  &\cc{3}26.75 &\cc{2}25.93 &\cc{1}24.58 &\cc{1}23.79 &\cc{2}0.905 &\cc{1}0.877 &\cc{1}0.816 &\cc{1}0.760 &\cc{3}0.100 &\cc{1}0.124 &\cc{1}0.227 &\cc{2}0.373 &\cc{1}0.063  \\
Ours (single ray)     &      25.38 &      25.72 &      24.27 &      23.43 &      0.881 &\cc{3}0.871 &      0.794 &      0.723 &      0.135 &      0.132 &      0.254 &      0.392 &      0.069  \\
Ours w/o scale        &\cc{2}26.81 &\cc{3}25.90 &\cc{3}24.52 &\cc{3}23.72 &\cc{3}0.904 &\cc{2}0.873 &\cc{3}0.807 &\cc{3}0.749 &\cc{2}0.098 &\cc{3}0.128 &\cc{3}0.235 &\cc{3}0.374 &\cc{2}0.064  \\
Ours w/o position     &      25.95 &      25.72 &      24.36 &      23.56 &      0.882 &      0.869 &      0.803 &      0.741 &      0.127 &      0.136 &      0.242 &\cc{2}0.373 &\cc{3}0.067  \\
Ours w/o patch        &\cc{1}26.85 &\cc{1}25.95 &\cc{2}24.56 &\cc{2}23.76 &\cc{1}0.906 &\cc{1}0.877 &\cc{2}0.813 &\cc{2}0.756 &\cc{1}0.097 &\cc{2}0.127 &\cc{2}0.231 &\cc{1}0.367 &\cc{2}0.064  \\
Ours w/o direction    &      26.35 &      25.36 &      24.10 &      23.38 &      0.899 &      0.864 &      0.799 &      0.744 &      0.103 &      0.137 &      0.242 &\cc{2}0.373 &\cc{3}0.067  \\
Ours w/o vis. weights &      25.90 &      25.11 &      23.91 &      23.19 &      0.888 &      0.856 &      0.792 &      0.737 &      0.118 &      0.148 &      0.250 &      0.378 &      0.070  \\
Ours (U-Net feat.)    &      26.00 &      25.10 &      23.43 &      22.77 &      0.887 &      0.853 &      0.761 &      0.706 &      0.116 &      0.151 &      0.298 &      0.427 &      0.075  \\

\hline
\end{tabularx}
\label{tb:multi-scale}
\end{table*}

\subsection{Datasets}
Our model is trained on three real datasets: real DTU multi-view dataset \cite{jensen2014large} and two real forward-facing datasets from LLFF \cite{mildenhall2019local} and IBRNet \cite{wang2021ibrnet}. All $190$ scenes (35 scenes from LLFF, 67 scenes from IBRNet and 88 scenes from DTU dataset) are used for training. Eight unseen scenes from the LLFF dataset are used for testing. During the multi-scale training, the resolutions of all input views are consistent, while the resolution of each target view is randomly selected from $1$ to $4$ times the input resolution. When training our model on single-scale datasets, the image resolutions of input and target images are the same. During testing, the resolution of input views is $504\times378$. We evaluate our model by rendering novel views at multiple scales: $\times 0.5$, $\times 1$, $\times 2$ and $\times 4$ ($\times 0.5$ denotes $0.5$ times the resolution of input views, and so on).

\subsection{Results}
We evaluate our model on LLFF testing scenes and compare it against three state-of-the-art generalizable methods: IBRNet \cite{wang2021ibrnet}, NeuRay \cite{liu2022neural} and GeoNeRF \cite{johari2022geonerf}. For quantitative evaluations, we report three error metrics, PSNR, SSIM and LPIPS \cite{zhang2018unreasonable}. Following prior work \cite{barron2021mip,suhail2022light}, we also summarize all three error metrics into an ``average" error metric by calculating the geometric mean of MSE$=10^{-\rm{PSNR}/10}$, $\sqrt{1-\rm{SSIM}}$, and LPIPS, which provides an easier comparison.

\noindent\textbf{Multi-scale novel views}. 
The performance of our method on rendering multi-scale novel views can be seen in \cref{tb:multi-scale}. and \cref{fig:results}. As shown in \cref{tb:multi-scale}, our method outperforms baseline methods on all four scales. Though our model isn't trained on $\times0.5$ scale, it also can render low-scale views by explicitly modifying the cone radius $r_m$ according to the pixel size at $\times0.5$ scale. In contrast, the baselines fail to render view at a lower scale due to the aliasing artifacts (shown as the $\times 0.5$ results by IBRNet in \cref{fig:teaser}(c)). We notice that the baselines produce relatively better results on rendering novel views at an up-scale after training on multi-scale datasets. On the other hand, they produce worse results on low-scale ($\times0.5$ and $\times1$) compared with their released models, since baselines have difficulty converging well when supervisions have different image scales, as discussed in \cref{fig:motivation} (b). Besides, our dataset contains fewer training scenes (about $15\%$ of the training scenes used by IBRNet or NeuRay). The qualitative comparisons are shown in \cref{fig:results}. It can be seen that the improvements produced by our method are most visually apparent in challenging cases such as small or thin structures. Besides, our renderings have fewer blurring artifacts and are closer to the ground truth. 

\begin{figure}[tb]
    \centering
    \includegraphics[width=\hsize]{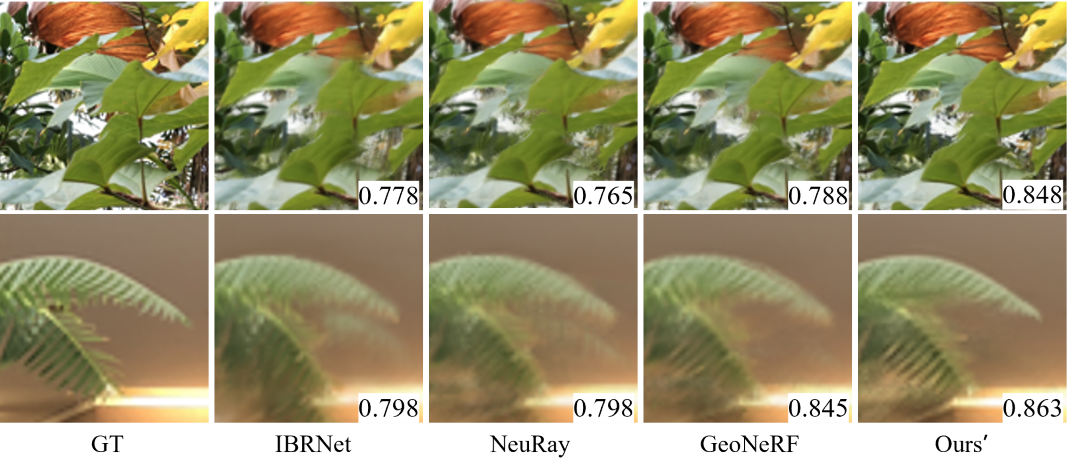}
    \vspace{-7mm}
    \caption{Qualitative comparisons on single-scale novel view synthesis. Though trained on the single-scale dataset, Ours$'$ still outperforms baselines. SSIM averaged over all novel views are provided at lower right. }
    \label{fig:single_scale}
\end{figure}

\noindent\textbf{Single-scale novel views.} 
We also train our model on the single-scale dataset to evaluate our ability on novel view synthesis. The quantitative results are shown in \cref{tb:single_scale}. One can see that training on the single-scale dataset does not reduce our performance on rendering novel views at $\times1$ scale, and even improves our results on PSNR and SSIM metrics. Figure \ref{fig:single_scale} presents the visualized comparisons on cropped regions. Although training on the single-scale dataset, our method shows a visual improvement on the challenging objects, such as the thin leaves.

\begin{figure}[tb]
    \centering
    \includegraphics[width=\hsize]{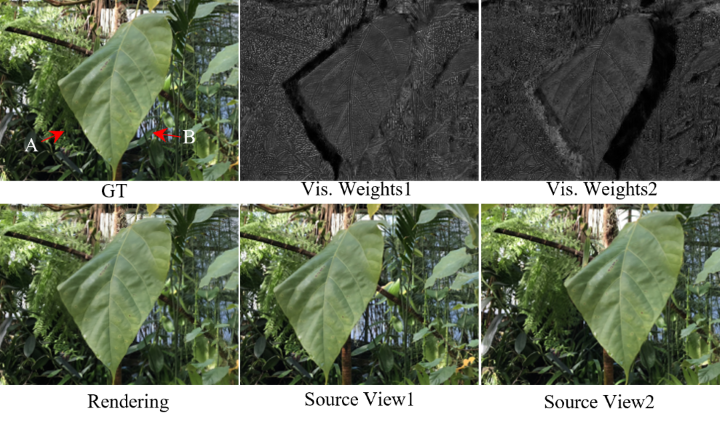}
    \vspace{-7mm}
    \caption{Visualizations of visibility weights (only two source views are presented). Our model accurately estimates the invisible regions for source views. For example, region A is occluded in source view 1, thus the visibility weights at this region are ``dark". }
    \label{fig:occlusion}
\end{figure}

\noindent\textbf{Visibility weights.} To demonstrate the effectiveness of our method in dealing with occlusions, we visualize the estimated visibility weights. As shown in \cref{fig:occlusion}, region A is occluded in source view 1, so it has smaller visibility weights in region A. Conversely, region A is visible in source view 2, while region B is occluded. Therefore, the visibility weights of source view 2 have larger values at region A and smaller values at region B. With visibility weights estimation, our method correctly renders the scene content at regions A and B.

\begin{table}[!tb]
\small
\centering
\caption{Quantitative comparisons of LIRF trained on single-scale dataset against baseline methods on rendering novel view at $\times 1$ scale. ``Ours$'$'' denotes our method training on single-scale dataset, while ``Ours'' denotes our method training on multi-scale dataset.}
\begin{tabularx}{\linewidth}{@{}lCCCCC}
\hline
                   & IBRNet & NeuRay & GeoNeRF  & Ours$'$ & Ours\\
\hline
PSNR$\uparrow$    & 25.28 & 25.17 & \cc{3}25.74 & \cc{1}26.30 & \cc{2}25.93 \\
SSIM$\uparrow$    & 0.840 & 0.837 & \cc{3}0.864 & \cc{1}0.878 & \cc{2}0.877 \\
LPIPS$\downarrow$ & 0.160 & 0.157 & \cc{3}0.136 & \cc{2}0.128 & \cc{1}0.124 \\
\hline
\end{tabularx}
\label{tb:single_scale}
\end{table}

\subsection{Ablation Studies} Ablation studies are shown in \cref{tb:multi-scale} to investigate the individual contribution of key modules of our model. (a) For ``single ray'', we modify our model to render a pixel from a single ray instead of a cone and keep everything else the same. Compared with our full model, the performance by ``single ray'' is reduced on all four testing scales, especially on the $\times0.5$ scale and $\times4$ scale, which demonstrates the contribution of our local implicit ray function for rendering multi-scale novel views. (b) For ``w/o scale'', we remove the scale $\mathbf{s}$ in \cref{eq:ray_function3} and \cref{eq:token1}. It reduces our model's ability to modify our implicit ray function according to the scale of target views.  (c) For ``w/o position'', the relative position $\Delta \mathbf{x}$ in \cref{eq:ray_function3} is removed, which prevents our implicit ray function from perceiving the relative 3D position between a sample and the vertices of conical frustum. (d) For ``w/o patch'', the size of the feature patches used to predict visibility weights is set to $1\times1$. This reduces our performance on rendering novel views with higher scales while improving our model on rendering $\times0.5$ novel views. When testing on different scales, the patch size is set to $7\times7$, which may be too large to render views at $\times0.5$ scale. (e) For ``w/o direction'', we remove the direction $\mathbf{d}$ in \cref{eq:token1}. Without considering the direction of source rays, our model produces worse results on all testing scales. (f) For ``w/o vis. weights'', we remove the visibility weights estimation module. Without visibility weights, our model fails to solve the occlusions, which significantly reduces our performance. (g) For ``U-Net feat.'', the image feature maps are extracted by the U-Net of IBRNet \cite{wang2021ibrnet}. We can see that our performance is reduced a lot. Most recent generalizable methods focus on the framework design of predicting color and densities from image features. Actually, the image feature extraction network is also important, affecting the quality of novel views from the source.

\section{Conclusion}
We have proposed a novel method for novel view synthesis of unseen scenes. Our method not only renders novel views with fewer blurring artifacts, but also produces novel views at arbitrary scales, even at higher scales than input views. The core of our method is the local implicit ray function that constructs each target ray by aggregating the information of a cone. Besides, our method also estimates the visibility weights to mitigate the occlusion problem. Compared with prior works, our method performs better on novel view synthesis of unseen scenes. Our code and models will be released to the research community to facilitate reproducible research.

\noindent\textbf{Limitations.}
Rendering novel views of unseen scenes is a challenging task. Similar to most image-based rendering methods, our method renders novel views from selected local input views, which causes two problems. First, our method can't render the regions occluded in all input views. Second, since selected input views vary with the novel view position, compared with NeRF, our method is weak in view consistency. Moreover, our method renders a pixel by casting four rays to approximate a cone, which increases the computational cost. Finally, though our method can mitigate the artifacts caused by occlusions, it fails in some challenging scenes. To solve this, a possible choice is introducing geometric priors.

\noindent\textbf{Acknowledgements.} 
The work was supported by NSFC under Grant 62031023.

{\small
\bibliographystyle{ieee_fullname}
\bibliography{egbib}
}

\clearpage
\section*{Supplemental Materials}
\setcounter{section}{0}
\renewcommand\thesection{\Alph{section}}
\section{Additional Implementation Details }
\subsection{Network Details}
Our feature extraction network (EDSR \cite{lim2017enhanced}) is based on the implementation from this URL (\url{https://github.com/sanghyun-son/EDSR-PyTorch}). The tail module of the EDSR network is removed, and two convolutional layers are added at the tail. The one outputs the feature maps for visibility weights, while the other one outputs the features maps for colors and densities. The implementation AE network is from GeoNeRF\cite{johari2022geonerf}. The ray function $\mathcal{R}$ is a three-layer MLP with 32 channels for each linear layer. Both networks $\mathcal{M}_w$ and $\mathcal{M}_\sigma$ are a two-layer MLP with 32 channels. The network $\mathcal{M}_c$ is a three-layer MLP with 32 channels. The transformer module $\mathcal{T}_1$ contains a single multi-head self-attention layer with the number of heads set to 4, while $\mathcal{T}_2$ contains four multi-head self-attention layers with the number of heads set to 4. The ``MLP'' used to reduce feature channels is a two-layer MLP and the number of channels is set to 32. For all MLP-based networks, ELU is used between each of two adjacent linear layers as the non-linear activation function. In our experiments, all networks are trained from scratch. Our code and model will be made available.

\subsection{Dataset Details}
We train our model on three real datasets: the real DTU multi-view dataset \cite{jensen2014large} and two real forward-facing datasets from LLFF \cite{mildenhall2019local} and IBRNet \cite{wang2021ibrnet}. All $190$ scenes (35 scenes from LLFF dataset, 67 scenes from IBRNet dataset and 88 scenes from DTU dataset) are used for training. We exclude the views with incorrect exposure from the DTU dataset as done in pixelNeRF \cite{yu2021pixelnerf}. Eight unseen scenes from LLFF dataset are used as our testing scenes. During the multi-scale training, the resolutions of all input views are consistent ($252\times189$ for LLFF and IBRNet datasets, $200\times150$ for DTU dataset), while the resolution of each target view is randomly selected from $1$ to $4$ times the input resolution (from $252\times189$ to $1008\times756$ for LLFF and IBRNet datasets, from $200\times150$ to $800\times600$ for DTU dataset). When training our model on single-scale datasets, the image resolutions of input and target images are the same ($504\times378$). During testing, the resolution of input views is $504\times378$. We evaluate our model on rendering novel views at multiple scales: $\times 0.5$, $\times 1$, $\times 2$ and $\times 4$ ($\times 0.5$ denotes $0.5$ times the resolution of input views, and so on). During the dataset preprocessing, we use bicubic interpolation to downsample high resolution images.

\begin{table}[!tb]
\small
\centering
\caption{ Fine-tuning results of our method and state-of-the-art methods. We fine-tune our pretrained model on each scene for 10k iterations with resolution of $1008\times756$. The resolution of testing views is also set to $1008\times756$. }
\vspace{-2mm}
\begin{tabularx}{\linewidth}{@{}l|C|C|C}
\hline
                   & PSNR$\uparrow$ &SSIM$\uparrow$  & LPIPS$\downarrow$ \\
\hline
NeRF\cite{mildenhall2020nerf}      &       26.50 &       0.811 & 0.250 \\
IBRNet\cite{wang2021ibrnet}    & \cc{3}26.73 & \cc{3}0.851 & 0.175 \\
NeuRay\cite{liu2022neural}    & \cc{1}27.06 &       0.850 & \cc{3}0.172 \\
GeoNeRF\cite{johari2022geonerf}   &       26.58 & \cc{2}0.856 & \cc{2}0.162 \\
Ours(10k) & \cc{2}26.85 & \cc{1}0.865 & \cc{1}0.159 \\
\hline
\end{tabularx}
\vspace{-3mm}
\label{tb:finetune}
\end{table}

\begin{table*}[!tb]
\small
\centering
\caption{Quantitative comparisons of varying the number of source views on LLFF real forward-facing scenes.}
\begin{tabularx}{\linewidth}{@{}l|CCCC|CCCC|CCCC|C}
\hline
       & \multicolumn{4}{c}{PSNR$\uparrow$} & \multicolumn{4}{|c}{SSIM$\uparrow$}  & \multicolumn{4}{|c|}{LPIPS$\downarrow$} & \multirow{2}{*}{Avg.$\downarrow$}\\
         \cline{2-5} \cline{6-9} \cline{10-13} 
       & $\times0.5$  & $\times1$ & $\times2$  & $\times4$  & $\times0.5$  & $\times1$ & $\times2$  & $\times4$ & $\times0.5$  & $\times1$ & $\times2$  & $\times4$  &\\
\hline
4 views  & \cc{0}24.80 & \cc{0}24.03 & \cc{0}22.93 & \cc{0}22.31 & \cc{0}0.864 & \cc{3}0.825 & \cc{0}0.761 & \cc{3}0.711 & \cc{0}0.134 & \cc{0}0.168 & \cc{0}0.269 & \cc{0}0.403 & \cc{0}0.080 \\
6 views  & \cc{3}26.11 & \cc{3}25.51 & \cc{3}24.24 & \cc{3}23.50 & \cc{3}0.893 & \cc{2}0.866 & \cc{3}0.804 & \cc{2}0.750 & \cc{3}0.106 & \cc{3}0.131 & \cc{3}0.233 & \cc{3}0.377 & \cc{3}0.066 \\
8 views  & \cc{1}26.75 & \cc{1}25.93 & \cc{1}24.58 & \cc{1}23.79 & \cc{1}0.905 & \cc{1}0.877 & \cc{1}0.816 & \cc{1}0.760 & \cc{1}0.100 & \cc{1}0.124 & \cc{1}0.227 & \cc{1}0.373 & \cc{1}0.063 \\
10 views & \cc{2}26.46 & \cc{2}25.91 & \cc{2}24.56 & \cc{2}23.78 & \cc{2}0.900 & \cc{1}0.877 & \cc{2}0.815 & \cc{1}0.760 & \cc{2}0.101 & \cc{2}0.126 & \cc{2}0.231 & \cc{2}0.376 & \cc{2}0.064 \\
\hline
\end{tabularx}
\label{tb:views}
\end{table*}

\begin{table}[!tb]
\footnotesize
\centering
\caption{Quantitative comparisons of our LIRF against two-staget methods on rendering novel view at higher scales ($\times2$ and $\times4$). We upsample low resolution novel views via bicubic interpolation (BI) or LIIF \cite{chen2021learning}. ``\textrm{M}" denotes the number of vertices used to represent a conical frustum. }
\begin{tabularx}{\linewidth}{@{}l|CC|CC|CC|C}
\hline
                   & \multicolumn{2}{c}{PSNR$\uparrow$} & \multicolumn{2}{|c}{SSIM$\uparrow$}  & \multicolumn{2}{|c|}{LPIPS$\downarrow$} & \multirow{2}{*}{Avg. $\downarrow$}\\
                   \cline{2-3} \cline{4-5} \cline{6-7}
                   & $\times2$  & $\times4$ & $\times2$  & $\times4$  & $\times2$  & $\times4$ & \\
\hline
IBRNet-BI    & 23.50 & 22.85 & 0.740 & 0.691 & 0.307 & 0.438 & 0.099 \\
IBRNet-LIIF  & 23.80 & 23.11 & 0.760 & 0.712 & 0.278 & 0.421 & 0.093 \\
NeuRay-BI    & 23.44 & 22.79 & 0.738 & 0.689 & 0.305 & 0.437 & 0.099 \\
NeuRay-LIIF  & 23.70 & 23.02 & 0.757 & 0.709 & 0.276 & 0.419 & 0.094 \\
GeoNeRF-BI   & 23.89 & 23.19 & 0.765 & 0.708 & 0.282 & 0.420 & 0.093 \\
GeoNeRF-LIIF & \cc{3}24.26 & \cc{3}23.53 & \cc{0}0.788 & \cc{0}0.733 &  0.251 & 0.400 & \cc{3}0.087 \\
\hline
Ours (M=4)        & \cc{0}23.91 &  23.15 & \cc{3}0.789 & \cc{3}0.741 & \cc{3}0.248 & \cc{3}0.398 & 0.089 \\
Ours (M=8)        & \cc{2}24.58 & \cc{2}23.79 & \cc{2}0.816 & \cc{2}0.760 & \cc{2}0.227 & \cc{2}0.373 & \cc{2}0.081 \\
Ours (M=10)        & \cc{1}24.93 & \cc{1}23.95 & \cc{1}0.838 & \cc{1}0.784 & \cc{1}0.218 & \cc{1}0.366 & \cc{1}0.077 \\
\hline
\end{tabularx}
\label{tb:two_stage}
\end{table}

\section{Additional Experiments}

\subsection{Fine-tuning}
Although our approach focuses on generalizations to unseen scenes, we also fine-turn our pre-trained model on each testing scene for comparison against previous methods. We follow the setting of IBRNet\cite{wang2021ibrnet} and train our model on each of the eight testing scenes for 10k iterations. The resolution of images used for training and testing is set to $1008 \times 756$. Note that the multi-view images used for fine-tuning are single-scale. The results are reported in \cref{tb:finetune}.

\subsection{Comparisons with Two-stage Methods} 
To further evaluate our method on rendering novel views at high scales ($\times2$ and $\times4$), we try to compare our method with two-stage methods. We first render novel views at $\times1$ scale via three baselines and then upsample the novel views via bicubic interpolation and a single-image super resolution method, LIIF \cite{chen2021learning}. The results are presented in \cref{tb:two_stage}. It shows the superiority of our model on rendering novel views at high scales with respect to the two-stage methods, though they introduce external data priors.

\subsection{Number of Source Views}
To investigate the robustness of our model to the number of source views, our model is tested on unseen scenes with different numbers of source views (4, 6, 8, and 10). The quantitative results are shown in \cref{tb:views}. The results show that our model produces competitive results when the number of source views is set to 6, 8, and 10. The model produces the best results when setting the number of source views to 8, since our model is trained with 8 source views. However, the performance of our method reduces a lot when the source views are sparse (4 views), since it is challenging to estimate visibility weights by matching sparse local image features.

\begin{figure}[tb]
    \centering
    \includegraphics[width=\hsize]{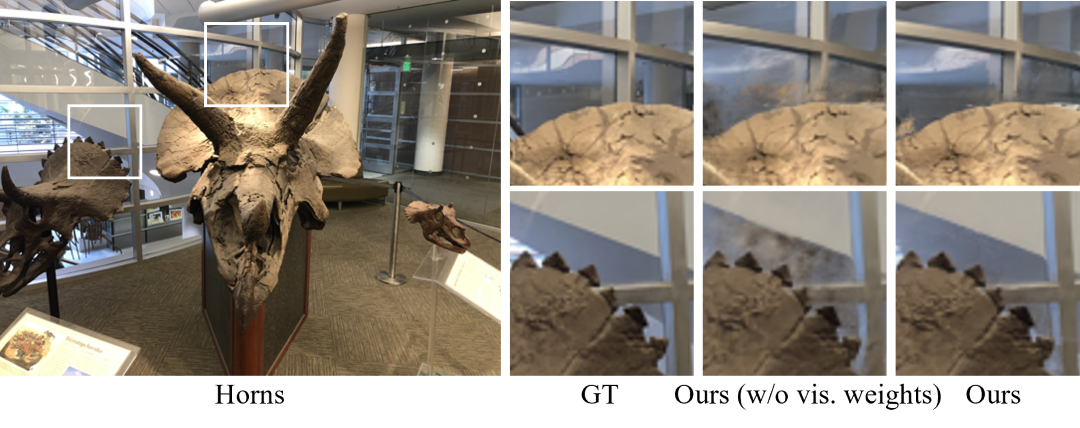}
    \vspace{-7mm}
    \caption{The qualitative results of our model without visibility weights. Ours denotes our full model.}
    \label{fig:wo_visw}
\end{figure}

\subsection{Number of Vertices}
We represent a conical frustum using several vertices. The results with different numbers of vertices are shown in \cref{tb:two_stage}. Using more vertices, our performance improves, but the rendering time increases too. Considering computing burdens and inspired by the voxel-based volume rendering, we use $\textrm{M}$ = 8 vertices to approximate a conical frustum. The samples within the conical frustum can be calculated by our implicit ray function.

\subsection{Comparisons of Rendering Time}
LIRF (45s for rendering an image with $\times 1$ scale ) is about three times slower than IBRnet ( 15s for rendering an image with $\times 1$ scale ). However, once the conical frustums are constructed, we directly infer rays from the conical frustums to render multi-scale views. Compared with baselines on rendering multi-scale views, we save the time of querying features from feature maps, especially on rendering high resolution views. 

\section{Additional Results}
\subsection{Qualitative Results for Ablation Studies}
\begin{figure*}[tb]
    \centering
    \includegraphics[width=\hsize]{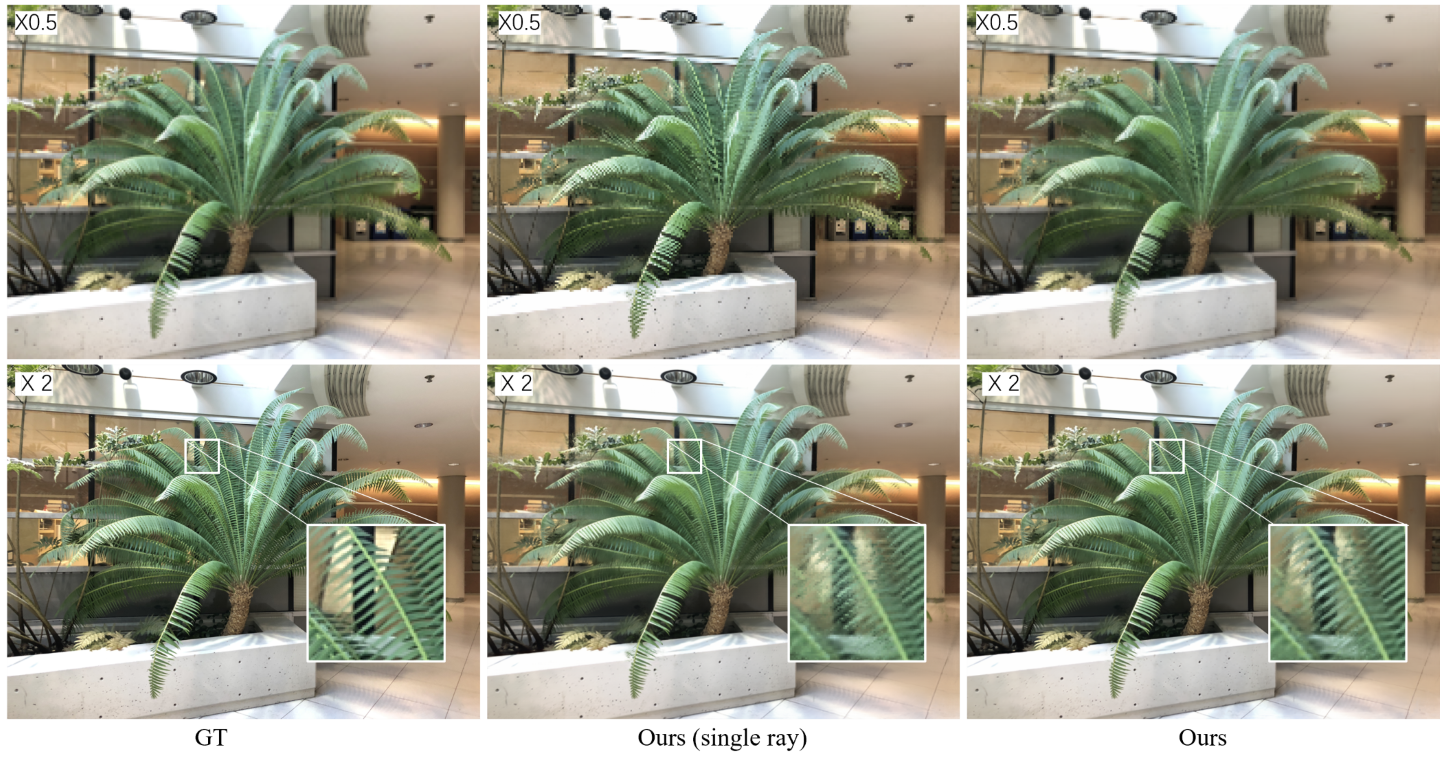}
    \vspace{-7mm}
    \caption{The qualitative results of our model that renders a pixel from a single ray. The top row shows the  novel views rendered at $\times 0.5$ scale. Our model (single ray) produces aliased novel view. The bottom row shows the novel views rendered at $\times 2$ scale. Our model (single ray) produces novel view with artifacts at thin structures. }
    \label{fig:single_ray}
\end{figure*}

\begin{figure*}[tb]
    \centering
    \includegraphics[width=\hsize]{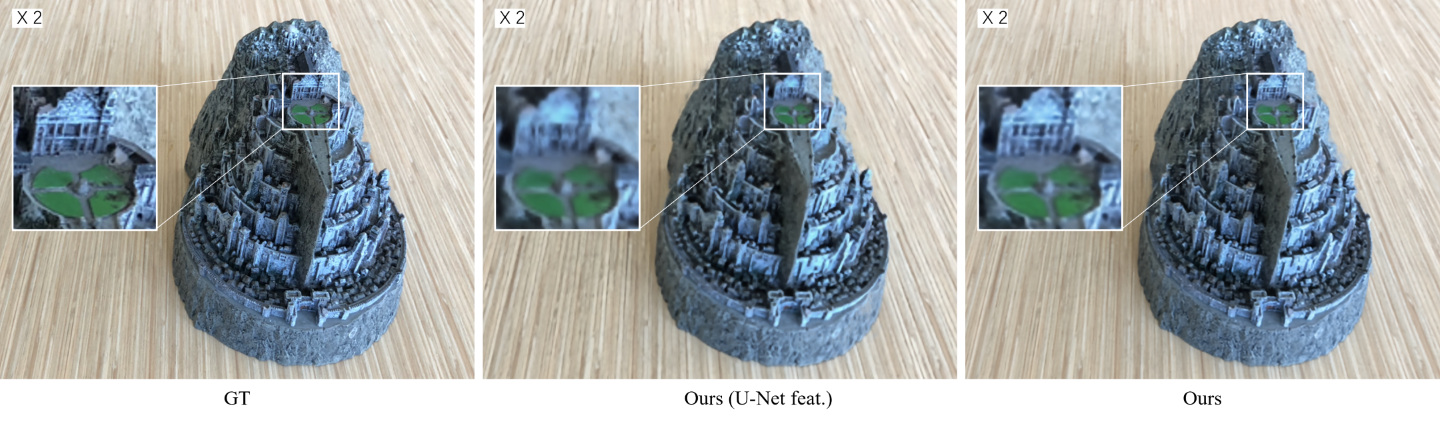}
    \vspace{-7mm}
    \caption{The qualitative results of our model that extracts image features via the U-Net in IBRNet \cite{wang2021ibrnet}. Our model (U-Net feat.) produces novel views with more blurred artifacts when the image features are extracted by the U-Net. }
    \label{fig:unet_feat}
\end{figure*}

As shown in \cref{tb:perscene}, three ablations (Ours(single ray), Ours w/o vis. weights and Ours(U-Net feat.)) mainly affect the performance of our LIRF. To further investigate their contributions to our model, the qualitative results are shown in \cref{fig:single_ray,fig:wo_visw,fig:unet_feat}.

\noindent\textbf{Ours w/o vis. weights.} We remove the visibility weights estimation module to evaluate the impact of the visibility weights. Figure \ref{fig:wo_visw} shows the performance of our model without  visibility weights. Our method produces renderings with ghosting artifacts on the boundary of objects due to occlusions. 

\noindent\textbf{Ours (single ray).} To investigate the contribution of our local implicit ray function, we render a pixel from a single ray instead of conical frustums. The results are presented in \cref{fig:single_ray}. One can see that our model (single ray) produces renderings that are excessively aliased when rendering novel views at $\times 0.5$ scale. Besides, our model (single ray) produces renderings containing artifacts at thin structures when rendering novel views at $\times 2$ scale.

\noindent\textbf{Ours (U-Net feat.).} Moreover, the feature extraction network is also important to our method, especially on rendering novel views at high scales. We therefore extract 2D image features via the U-Net in IBRNet \cite{wang2021ibrnet}. Our model with the U-Net is trained from scratch on our multi-scale dataset. The rendered testing views are presented in \cref{fig:unet_feat}. Our model produces renderings with more blurred artifacts when the image features are extracted by the U-Net. 

\subsection{A Failure Case}
\begin{figure}[tb]
    \centering
    \includegraphics[width=\linewidth]{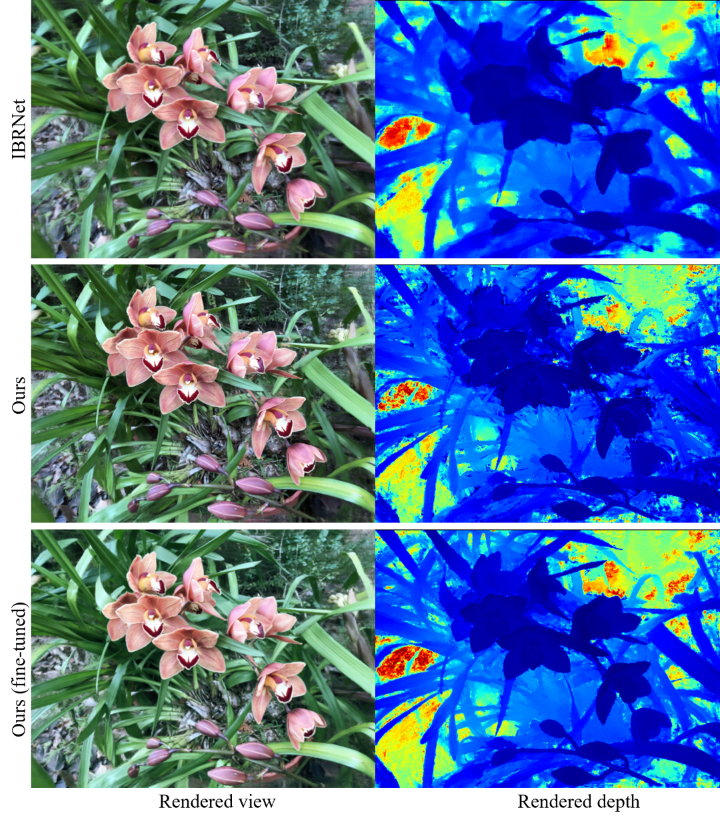}
    \vspace{-7mm}
    \caption{A failure example on \textit{orchids} scene. Our model fails to predict the geometry on the boundaries of flowers due to occlusions. The renderings by IBRNet \cite{wang2021ibrnet} also contain blurred artifacts. After fine-tuning on this scene for 10k iterations, our model produces results with fewer artifacts}
    \label{fig:limitations}
\end{figure}

As discussed in the limitations, though the visibility weights can mitigate the artifacts caused by occlusions, they fail in some challenging scenes such as the \textit{orchids} scene. Figure \ref{fig:limitations} shows a failure example on the \textit{orchids} scene. The multi-view images of this scene are captured sparsely, which is challenging for our model to estimate the accurate visibility weights. The baselines also struggle with this challenging scene, such as the renderings by IBRNet \cite{wang2021ibrnet} with blurred artifacts. After fine-tuning on this scene for 10k iterations, our model produces results with fewer artifacts on the boundary of objects.

\subsection{Per-Scene Results}
To evaluate our approach compared to previous methods on each individual scene, per-scene results on the eight testing scenes are presented in \cref{tb:perscene}. We report the arithmetic mean of each metric averaged over the four testing scales used for testing. Our method yields a significant improvement in three error metrics across most scenes.

\begin{table*}[!tb]
\small
\centering
\caption{Per scene quantitative comparisons of our LIRF and its ablations against IBRNet \cite{wang2021ibrnet}, NeuRay \cite{liu2022neural} and GeoNeRF \cite{johari2022geonerf} on LLFF\cite{mildenhall2019local} multi-scale testing dataset. Metrics are averaged over four testing scales ($\times 0.5$, $\times 1$, $\times 2$ and $\times 4$). $*$ denotes training on the same multi-scale training set as our method.}
\begin{tabularx}{\linewidth}{@{}l|CCCCCCCC}
\hline
& \multicolumn{8}{c}{Average PSNR$\uparrow$} \\
\cline{2-9}
& \textit{fern}  & \textit{flower} & \textit{fortress}  & \textit{horns} & \textit{leaves}  & \textit{orchids} & \textit{room}  & \textit{trex} \\
\hline
IBRNet                & \cc{0}23.40 & \cc{0}25.80 & \cc{0}28.12 & \cc{0}24.78 & \cc{0}19.69 & \cc{0}19.20 & \cc{0}27.49 & \cc{0}22.89 \\
NeuRay                & \cc{0}23.21 & \cc{0}25.89 & \cc{0}28.18 & \cc{0}24.78 & \cc{0}19.48 & \cc{0}18.87 & \cc{0}27.09 & \cc{0}22.81 \\
GeoNeRF               & \cc{0}23.73 & \cc{0}26.35 & \cc{0}28.88 & \cc{0}25.19 & \cc{0}19.75 & \cc{1}19.81 & \cc{0}27.02 & \cc{0}21.91 \\
IBRNet*               & \cc{0}22.38 & \cc{0}24.61 & \cc{0}26.32 & \cc{0}23.68 & \cc{0}18.48 & \cc{0}18.07 & \cc{0}25.49 & \cc{0}21.80 \\
NeuRay*               & \cc{0}21.26 & \cc{0}23.56 & \cc{0}25.43 & \cc{0}22.45 & \cc{0}17.89 & \cc{0}17.50 & \cc{0}25.12 & \cc{0}20.84 \\
GeoNeRF*              & \cc{0}23.55 & \cc{0}26.21 & \cc{0}28.17 & \cc{0}24.92 & \cc{0}19.89 & \cc{2}19.50 & \cc{0}26.24 & \cc{0}21.47 \\
\hline
Ours                  & \cc{1}25.21 & \cc{1}26.77 & \cc{3}29.07 & \cc{1}26.59 & \cc{1}21.59 & \cc{0}19.39 & \cc{0}28.59 & \cc{3}24.91 \\
Ours w/o scale        & \cc{3}25.05 & \cc{0}26.63 & \cc{1}29.25 & \cc{2}26.56 & \cc{3}21.33 & \cc{0}19.28 & \cc{1}28.75 & \cc{2}25.06 \\
Ours w/o patch        & \cc{2}25.18 & \cc{3}26.64 & \cc{2}29.17 & \cc{3}26.47 & \cc{1}21.47 & \cc{3}19.44 & \cc{2}28.72 & \cc{1}25.16 \\
Ours w/o position     & \cc{0}24.78 & \cc{2}26.69 & \cc{0}28.18 & \cc{0}26.16 & \cc{0}20.97 & \cc{0}19.33 & \cc{0}28.42 & \cc{0}24.64 \\
Ours w/o direction    & \cc{0}24.60 & \cc{0}26.31 & \cc{0}28.34 & \cc{0}25.65 & \cc{0}20.86 & \cc{0}19.22 & \cc{3}28.66 & \cc{0}24.73 \\
Ours w/o vis. weights & \cc{0}24.72 & \cc{0}25.95 & \cc{0}27.92 & \cc{0}25.50 & \cc{0}20.66 & \cc{0}18.98 & \cc{0}27.75 & \cc{0}24.75 \\
Ours (U-Net feat.)    & \cc{0}24.21 & \cc{0}26.03 & \cc{0}28.67 & \cc{0}25.33 & \cc{0}20.44 & \cc{0}19.14 & \cc{0}27.12 & \cc{0}23.67 \\
Ours (single ray)     & \cc{0}24.49 & \cc{0}26.60 & \cc{0}28.11 & \cc{0}25.78 & \cc{0}20.83 & \cc{0}19.28 & \cc{0}28.07 & \cc{0}24.41 \\
\hline
\end{tabularx}

\vspace{5mm}

\begin{tabularx}{\linewidth}{@{}l|CCCCCCCC}
\hline
& \multicolumn{8}{c}{Average SSIM$\uparrow$} \\
\cline{2-9}
& \textit{fern}  & \textit{flower} & \textit{fortress}  & \textit{horns} & \textit{leaves}  & \textit{orchids} & \textit{room}  & \textit{trex} \\
\hline
IBRNet                & \cc{0}0.741 & \cc{0}0.836 & \cc{0}0.832 & \cc{0}0.805 & \cc{0}0.678 & \cc{0}0.629 & \cc{0}0.899 & \cc{0}0.794 \\
NeuRay                & \cc{0}0.739 & \cc{0}0.836 & \cc{0}0.833 & \cc{0}0.808 & \cc{0}0.668 & \cc{0}0.617 & \cc{0}0.896 & \cc{0}0.790 \\
GeoNeRF               & \cc{0}0.768 & \cc{0}0.847 & \cc{0}0.844 & \cc{0}0.825 & \cc{0}0.683 & \cc{3}0.659 & \cc{0}0.897 & \cc{0}0.795 \\
IBRNet*               & \cc{0}0.717 & \cc{0}0.820 & \cc{0}0.801 & \cc{0}0.790 & \cc{0}0.650 & \cc{0}0.593 & \cc{0}0.877 & \cc{0}0.786 \\
NeuRay*               & \cc{0}0.675 & \cc{0}0.761 & \cc{0}0.723 & \cc{0}0.733 & \cc{0}0.568 & \cc{0}0.531 & \cc{0}0.862 & \cc{0}0.735 \\
GeoNeRF*              & \cc{0}0.774 & \cc{0}0.852 & \cc{0}0.833 & \cc{0}0.829 & \cc{0}0.704 & \cc{0}0.655 & \cc{0}0.893 & \cc{0}0.802 \\
\hline
Ours                  & \cc{1}0.825 & \cc{1}0.870 & \cc{2}0.897 & \cc{1}0.876 & \cc{1}0.787 & \cc{2}0.666 & \cc{1}0.924 & \cc{1}0.872 \\
Ours w/o scale        & \cc{3}0.817 & \cc{3}0.865 & \cc{3}0.896 & \cc3{}0.870 & \cc{1}0.776 & \cc{0}0.656 & \cc{3}0.921 & \cc{3}0.866 \\
Ours w/o patch        & \cc{2}0.821 & \cc{2}0.867 & \cc{1}0.898 & \cc{2}0.875 & \cc{2}0.782 & \cc{1}0.668 & \cc{2}0.923 & \cc{2}0.870 \\
Ours w/o position     & \cc{0}0.806 & \cc{0}0.858 & \cc{0}0.882 & \cc{0}0.863 & \cc{0}0.761 & \cc{0}0.652 & \cc{0}0.913 & \cc{0}0.856 \\
Ours w/o direction    & \cc{0}0.806 & \cc{0}0.861 & \cc{0}0.891 & \cc{0}0.864 & \cc{0}0.756 & \cc{0}0.651 & \cc{0}0.920 & \cc{0}0.864 \\
Ours w/o vis. weights & \cc{0}0.807 & \cc{0}0.849 & \cc{0}0.886 & \cc{0}0.853 & \cc{0}0.748 & \cc{0}0.635 & \cc{0}0.910 & \cc{0}0.860 \\
Ours (U-Net feat.)    & \cc{0}0.777 & \cc{0}0.849 & \cc{0}0.858 & \cc{0}0.831 & \cc{0}0.728 & \cc{0}0.642 & \cc{0}0.898 & \cc{0}0.829 \\
Ours (single ray)     & \cc{0}0.799 & \cc{0}0.856 & \cc{0}0.873 & \cc{0}0.849 & \cc{0}0.757 & \cc{0}0.651 & \cc{0}0.904 & \cc{0}0.851 \\
\hline
\end{tabularx}

\vspace{5mm}

\begin{tabularx}{\linewidth}{@{}l|CCCCCCCC}
\hline
& \multicolumn{8}{c}{Average LPIPS$\downarrow$} \\
\cline{2-9}
& \textit{fern}  & \textit{flower} & \textit{fortress}  & \textit{horns} & \textit{leaves}  & \textit{orchids} & \textit{room}  & \textit{trex} \\
\hline
IBRNet                & \cc{0}0.282 & \cc{0}0.201 & \cc{0}0.195 & \cc{0}0.252 & \cc{0}0.285 & \cc{0}0.316 & \cc{0}0.214 & \cc{0}0.272 \\
NeuRay                & \cc{0}0.282 & \cc{0}0.191 & \cc{0}0.189 & \cc{0}0.246 & \cc{0}0.293 & \cc{0}0.311 & \cc{0}0.206 & \cc{0}0.265 \\
GeoNeRF               & \cc{0}0.251 & \cc{0}0.187 & \cc{0}0.170 & \cc{0}0.226 & \cc{0}0.283 & \cc{1}0.287 & \cc{0}0.207 & \cc{0}0.267 \\
IBRNet*               & \cc{0}0.297 & \cc{0}0.208 & \cc{0}0.221 & \cc{0}0.263 & \cc{0}0.297 & \cc{0}0.339 & \cc{0}0.235 & \cc{0}0.279 \\
NeuRay*               & \cc{0}0.359 & \cc{0}0.269 & \cc{0}0.296 & \cc{0}0.336 & \cc{0}0.369 & \cc{0}0.395 & \cc{0}0.262 & \cc{0}0.331 \\
GeoNeRF*              & \cc{0}0.245 & \cc{3}0.176 & \cc{0}0.181 & \cc{0}0.224 & \cc{0}0.264 & \cc{2}0.288 & \cc{0}0.212 & \cc{0}0.265 \\
\hline
Ours                  & \cc{1}0.217 & \cc{1}0.174 & \cc{2}0.152 & \cc{2}0.191 & \cc{1}0.219 & \cc{2}0.288 & \cc{0}0.190 & \cc{2}0.219 \\
Ours w/o scale        & \cc{3}0.223 & \cc{0}0.177 & \cc{1}0.149 & \cc{3}0.193 & \cc{3}0.226 & \cc{0}0.296 & \cc{3}0.188 & \cc{3}0.220 \\
Ours w/o patch        & \cc{2}0.221 & \cc{2}0.175 & \cc{1}0.149 & \cc{1}0.187 & \cc{2}0.222 & \cc{3}0.289 & \cc{1}0.185 & \cc{1}0.216 \\
Ours w/o position     & \cc{0}0.234 & \cc{0}0.181 & \cc{0}0.165 & \cc{0}0.199 & \cc{0}0.254 & \cc{0}0.311 & \cc{3}0.188 & \cc{0}0.223 \\
Ours w/o direction    & \cc{0}0.231 & \cc{0}0.182 & \cc{3}0.153 & \cc{0}0.196 & \cc{0}0.233 & \cc{0}0.303 & \cc{2}0.187 & \cc{0}0.223 \\
Ours w/o vis. weights & \cc{0}0.233 & \cc{0}0.193 & \cc{0}0.162 & \cc{0}0.209 & \cc{0}0.247 & \cc{0}0.320 & \cc{0}0.199 & \cc{0}0.225 \\
Ours (U-Net feat.)    & \cc{0}0.266 & \cc{0}0.199 & \cc{0}0.200 & \cc{0}0.246 & \cc{0}0.269 & \cc{0}0.313 & \cc{0}0.227 & \cc{0}0.263 \\
Ours (single ray)     & \cc{0}0.240 & \cc{0}0.188 & \cc{0}0.181 & \cc{0}0.214 & \cc{0}0.257 & \cc{0}0.318 & \cc{0}0.200 & \cc{0}0.229 \\
\hline
\end{tabularx}
\label{tb:perscene}
\end{table*}

\end{document}